\documentclass[conference]{IEEEtran}
\IEEEoverridecommandlockouts

\usepackage{cite}
\usepackage{amsmath,amssymb,amsfonts}
\usepackage{algorithmic}
\usepackage{graphicx}
\usepackage{textcomp}
\usepackage{graphicx}
\usepackage[export]{adjustbox}
\usepackage{babel,blindtext}
\usepackage{xcolor}
\def\BibTeX{{\rm B\kern-.05em{\sc i\kern-.025em b}\kern-.08em
    T\kern-.1667em\lower.7ex\hbox{E}\kern-.125emX}}
\begin{document}

 \title {Object Detection for Vehicle Dashcams using Transformers}


 \author{\IEEEauthorblockN{F L}
\IEEEauthorblockA{\textit{Center } \\
\textit{Bahria University}\\
Islamabad, Pakistan \\
muhammadosama939@gmail.com}
\and
\IEEEauthorblockN{ Khizer Ali}
\IEEEauthorblockA{\textit{Center of Excellence in AI} \\
\textit{Bahria University}\\
Islamabad, Pakistan  \\
mkhizer.buic@bahria.edu.pk}
\and
\IEEEauthorblockN{Anam Bibi }
\IEEEauthorblockA{\textit{Center of Excellence in AI} \\
\textit{Bahria University}\\
Islamabad, Pakistan \\
anam.buho@bahria.edu.pk}

\and
\IEEEauthorblockN{Imran Siddiqi}
\IEEEauthorblockA{\textit{Center of Excellence in AI} \\
\textit{Bahria University}\\
Islamabad, Pakistan \\
imran.siddiqi@bahria.edu.pk}
}
\author{\IEEEauthorblockN{Osama Mustafa}
\IEEEauthorblockA{\textit{Center of Excellence in AI} \\
\textit{Bahria University}\\
Islamabad, Pakistan \\
muhammadosama939@gmail.com}
\and
\IEEEauthorblockN{ Khizer Ali}
\IEEEauthorblockA{\textit{Center of Excellence in AI} \\
\textit{Bahria University}\\
Islamabad, Pakistan  \\
mkhizer.buic@bahria.edu.pk}
\and
\IEEEauthorblockN{Anam Bibi }
\IEEEauthorblockA{\textit{Center of Excellence in AI} \\
\textit{Bahria University}\\
Islamabad, Pakistan \\
anam.buho@bahria.edu.pk}

\and
\IEEEauthorblockN{Imran Siddiqi}
\IEEEauthorblockA{\textit{Center of Excellence in AI} \\
\textit{Bahria University}\\
Islamabad, Pakistan \\
imran.siddiqi@bahria.edu.pk}

\and
\IEEEauthorblockN{Momina Moetesum}
\IEEEauthorblockA{\textit{Center of Excellence in AI} \\
\textit{Bahria University}\\
Islamabad, Pakistan \\
momina.buic@bahria.edu.pk}

}

\maketitle

\begin{abstract}
The use of intelligent automation is growing significantly in the automotive industry, as it assists drivers and fleet management companies, thus increasing their productivity. Dash cams are now been used for this purpose which enables the instant identification and understanding of multiple objects and occurrences in the surroundings. In this paper, we propose a novel approach for object detection in dashcams using transformers. Our system is based on the state-of-the-art DEtection TRansformer (DETR), which has demonstrated strong performance in a variety of conditions, including different weather and illumination scenarios. The use of transformers allows for the consideration of contextual information in decision-making, improving the accuracy of object detection. To validate our approach, we have trained our DETR model on a dataset that represents real-world conditions. Our results show that the use of intelligent automation through transformers can significantly enhance the capabilities of dashcam systems. The model achieves an mAP of 0.95 on detection.
\end{abstract}


\section{Introduction}
Dashcams are an important tool for increasing road safety and efficiency, as they enable real-time monitoring and analysis of the driving environment. One key aspect of intelligent truck dashcams is object detection, which involves the identification and classification of various objects and events in the environment. Accurate object detection is crucial for the proper functioning of intelligent truck dashcam systems, as it allows for the identification of potential hazards, the tracking of vehicles and pedestrians, and the recognition of traffic signs and signals. This object detection and classification is also an important step towards autonomous /self-driving.

In this paper, we present a novel approach for object detection in dashcams which make them intelligent using transformers. Transformers are a class of neural network architectures that have achieved state-of-the-art performance in many natural language processing tasks and have recently been applied to a variety of computer vision tasks as well \cite{vaswani2017attention,han2022survey}. We demonstrate the effectiveness of the DEtection TRansformer (DETR) \cite{carion2020end} for object detection in intelligent dashcams, and show that our system performs well in a variety of different conditions. In addition to describing our proposed approach, we also present experimental results that validate the effectiveness of our method. DETR has outperformed state of the art object detectors like YOLO and RCNN variants in other challenging scenarios like underwater object detection \cite{ali2022marine}.  

Object detection in vehicle dashcams is a challenging problem with the following challenges:

\begin{itemize}
  \item Highly dynamic environment on road. The traffic state on a road is continuously changing especially in long-route traveling vehicles such as Trucks
  \item Different illumination conditions due to different daytime, weather and scene
  \item Different challenging scenarios such as angle, orientation, occlusion and small size of stop signs
  \item Many of the object detectors perform well in training but performance drops in deployment conditions, so the detecting network is not able to generalize well on real-world conditions
  
\end{itemize}

The trucking industry is using intelligent automation in trucks for early warning and decision systems to prevent accidents. In this work, the trained object detection network DETR performs very well in challenging conditions even when a human driver would face difficulty in decision-making. The network has been trained on a dataset that has been collected in real-world conditions by deploying a dashcam on a fleet of trucks. This proposed solution can play an efficient role in the intelligent automation of truck dashcams. The main contributions of our work are as follows:
 \begin{itemize}
    \item Detection of vehicles and road signs on real-world dash-cam datasets that contains images from different challenging scenarios.
    \item Investigation of transformer-based object detection for dash-cam object detection. 
\end{itemize}

This paper is structured as follows: in Section~\ref{sec:lit}, we provide a summary of significant advances in vehicle object detection. The dataset used in our research is outlined in Section~\ref{sec:dataset}. The methodology is outlined in section~\ref{sec:method} Our experimental study, results and analysis are discussed in Section~\ref{sec:res}. Finally, the paper concludes in Section~\ref{sec:con} with highlights of our main findings.

\section{Related work}\label{sec:lit}
In autonomous driving, Traffic signs (Stop-signs), traffic signals and other object detection  is an important and challenging problems due to the illumination variations and background clutter. The importance of addressing these challenges, as autonomous driving has the potential to significantly improve the safety, efficiency, and accessibility of vehicles. Accurate perception and understanding of the environment are crucial for the vehicle to be able to navigate safely and avoid collisions. Real-time performance is also essential for the vehicle to respond to changing situations and events in a timely manner.
Most of the previous studies focused on the recognition or classification of traffic signs and other objects.

\subsection{Object Detection}
\begin{table*}[!h]
\centering
\caption{Summary of State-of-the-Arts}
\label{tab:my-table1}
\begin{tabular}{llll}
\hline
\hline
\textbf{Task} & \textbf{Dataset} & \textbf{Technique} & \textbf{Results} \\ 
\hline
Parking Sign detection \cite{jin2022real} & Custom Dataset & YOLOv5 &  0.96 mAP \\ 
Traffic Sign Recognition \cite{lau2015malaysia} & Malaysia Traffic Sign Dataset (MTSD) &  CNN & 0.99 RMSE\\ 
 Detection of Stop Sign Violations \cite{bravi2021detection} &  Custom dataset (dashcam video) &  YOLOv3 & 0.94 mAP \\ 
Traffic sign classification and localization \cite{houben2013detection} & German traffic sign detection dataset (GTSD)& - & 0.90 mAP\\ 
Traffic sign detection and recognition \cite{mykola_2018} & German traffic sign detection (GTSD) & YOLOv5 & 0.88 mAP \\ 
 Object Detection of traffic violations \cite{franklin2020traffic} & dashcam video & YOLOv3 & 89.2 Accuracy\\ 
 Vehicle detection  \cite{karthika2022novel} & GTSDB \& GTSRB dataset  & You only look twice (YOLT) & 89.2 mAP\\ 
Traffic lights and sign detection \cite{jayasinghe2022towards} &  CeyRo dataset & SSD & 0.92 F1 score\\ 
\hline 
\end{tabular}
\end{table*}
In computer vision, Object detection is a challenging problem and a highly active area of research. The goal of object detection is to determine the object's location and class within an image or video. In recent years, deep learning techniques are powerful for the representation of feature learning from object detection data directly and in the fields of generic object detection the deep learning techniques are huge (main) breakthroughs {\cite{liu2020deep}. The deep learning models are divided into two categories for the tasks of object localization and recognition namely Two-Stage and One-stage \cite{carranza2020performance}. Although the detection performance of two-stage detectors is good, their processing speed is slow and requires high computational costs. The One-Stage detector created a balance between accuracy and speed. However, in the last few years, the most popular Two-Stage and One-Stage models are Faster R-CNN \cite{girshick2015fast}, Mask RCNN \cite{he2017mask}, FPN\cite{lin2017feature}, SSD \cite{liu2016localization}, YOLO \cite{jiang2022review} for object detection. Recently, transformer-based techniques or methods have been used in various fields for object detection. In 2020, Carion et al. \cite{carion2020end} introduced a new method for object detection known as "Detection  Transformer" which was based on the transformer and bipartite matching loss with parallel decoding. The previous detector with RNNs used autoregressive decoding \cite{carion2020end}. Due to using parallel processing (Not using NMS and anchors boxes techniques) DeTr performs fast as compared to previous detectors. The proposed model DeTr performance was evaluated on MS-COCO (Large Dataset). In addition, the overall architecture of DeTr is simple and more powerful in the image where context is important as compared to other detectors.
\par In recent years, most of the studies used deep learning techniques for this problem but most of them have not been successful when applied in-field (real-world) environment. Due to the constrained environment database and small benchmarks. In \cite{chaureal} the authors introduced the database of street-based parking sign detection. They used different YOLOv5 models including YOLOv5s, YOLOv5m, and Swim Transformer. However, the proposed solution was based on the YOLOv5 model and achieved 96.8\%  accuracy. But the proposed solution sometimes failed when testing the model on dashcam video. In another research, Mian et al. presented a CNN-based solution and used the large Malaysia traffic sign database only for the recognition of traffic sign \cite{lau2015malaysia}.

\par In the last few years, several databases are introduced for traffic sign detection and recognition. The large real-world dataset is "German Traffic Sign Detection Benchmark" \cite{houben2013detection} presented in competition at IJCNN and used for the localization and classification of the traffic signs. In this dataset, the captured images contain a natural and illumination variation but it only used the cropped around the traffic sign images. However, most of the datasets contain cropped images of traffic lights and traffic signs, which have been extracted from tencent and google (other search engines) and most of the studies collected datasets using cameras footage mounted in vehicles such as \cite{zhang2022cctsdb}, \cite{stallkamp2012man}, \cite{mogelmose2012vision} and \cite{gray2022glare}.
\par Road accidents are often caused by the violation of stop signs in daily life. Bravi et al. developed an automatic system for stop sign violation detection. The proposed solution is based on the YOLOv3 model and the performance evaluated on the video dashcam dataset \cite{bravi2021detection}. In another study \cite{franklin2020traffic}, the traffic violation system was developed based on YOLOv3 using dashcam video. The object detection of traffic violations such as the number of vehicles, speed of vehicles and the jump signal. The proposed model obtained 89.2\% for the detection of traffic violations and achieved 97.6\% accuracy for the count of vehicle detection. Besides, for detection of road stop signs detection using driving data \cite{dadras2019novel} introduced a novel algorithm based on a statistical analysis of obtained drive history data. \\
\begin{figure*}[!ht]
\centering
\begin{tabular}{cc}
  \includegraphics[width=70mm]{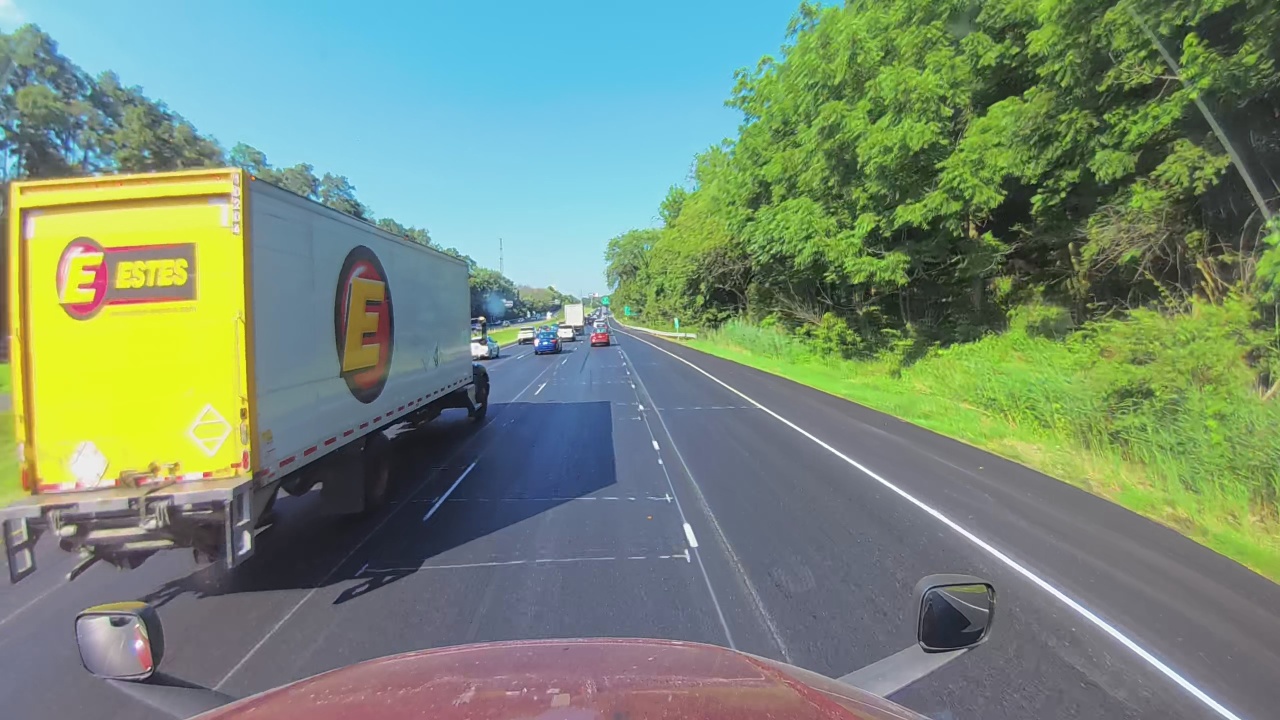} &   \includegraphics[width=70mm]{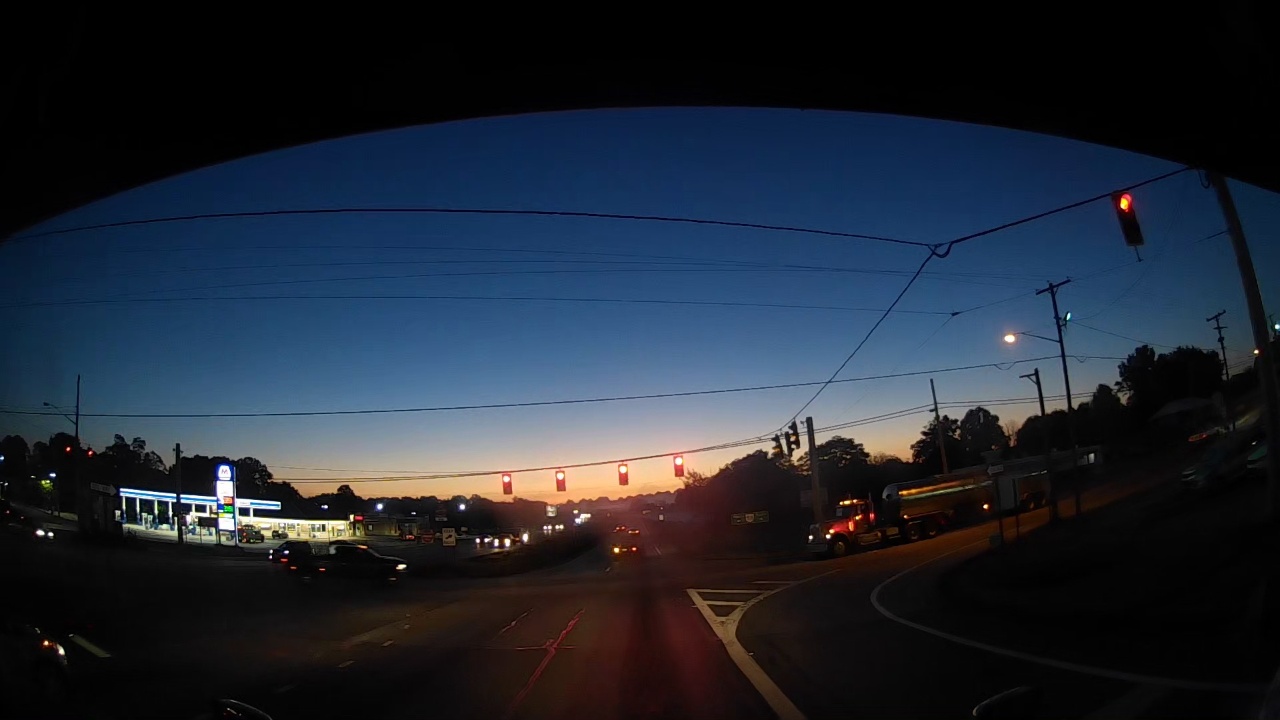} \\ 
 (a) & (b) \\[6pt]

\end{tabular}
\caption{Sample images of dataset}
\label{Fig:dataset}
\end{figure*}

\par As already mentioned, the main contribution of this study is to detect objects in challenging problems such as illumination variation, different weather conditions, and multi-scales objects (Small, medium, large) using truck dashcam data. More recently, In \cite{karthika2022novel} authors presented an efficient algorithm based on YOLOv3 for the improvement and enhancement of the performance of the Advanced Driver assistance system (ADAS). In this study, they addressed the real-time condition challenges. But they used the large old German Traffic Sign Detection Benchmark (GTSDB) \cite{mykola_2018} for the detection and recognition of traffic signs for self-driving. GTSDB database contains the cropped traffic sign images although only the traffic sign cropped images detect and recognized, however, in real-time we have multiple challenges regarding the resolution of images. GTSDB is also an unbalanced dataset. However, they used two separate models: for the detection they achieved 89.9\% accuracy and for the classification (Recognition) achieved 86\% accuracy. To address, the problem of challenging road scenarios such as weather conditions and illumination Jayasinghe et al.\cite{jayasinghe2022towards} introduced an end-to-end and simple detection framework for the traffic light and traffic sign detection. The proposed solution detects the traffic light and signs in complex road scenarios and the solution is based on a Two-stage SSD detector. Additionally, They introduced a new dataset known as "CeyRo" and it contains 7.9k images based on 75 classes of traffic lights and traffic signs. Due to the success of the large and in-field challenges dataset, we have employed the same in our study.

\begin{figure}[!h]
    
    \includegraphics[width = 0.5\textwidth]{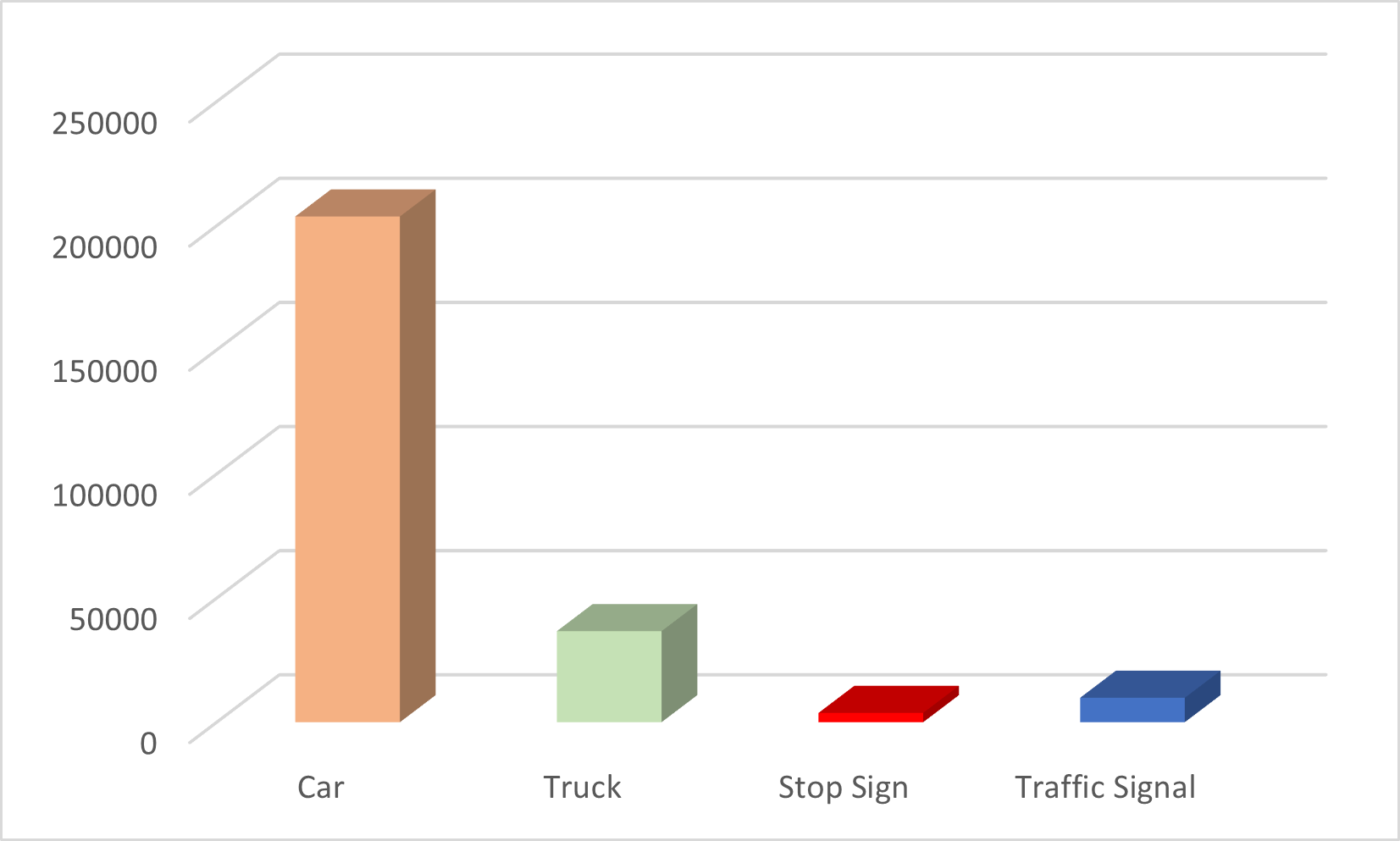}
    \caption{Histogram representing class distribution in dataset}
    \label{fig:hist}
\end{figure}

\begin{figure*}[!ht]

   \includegraphics[trim={0.2 0 0 0.2},width=0.9\linewidth]{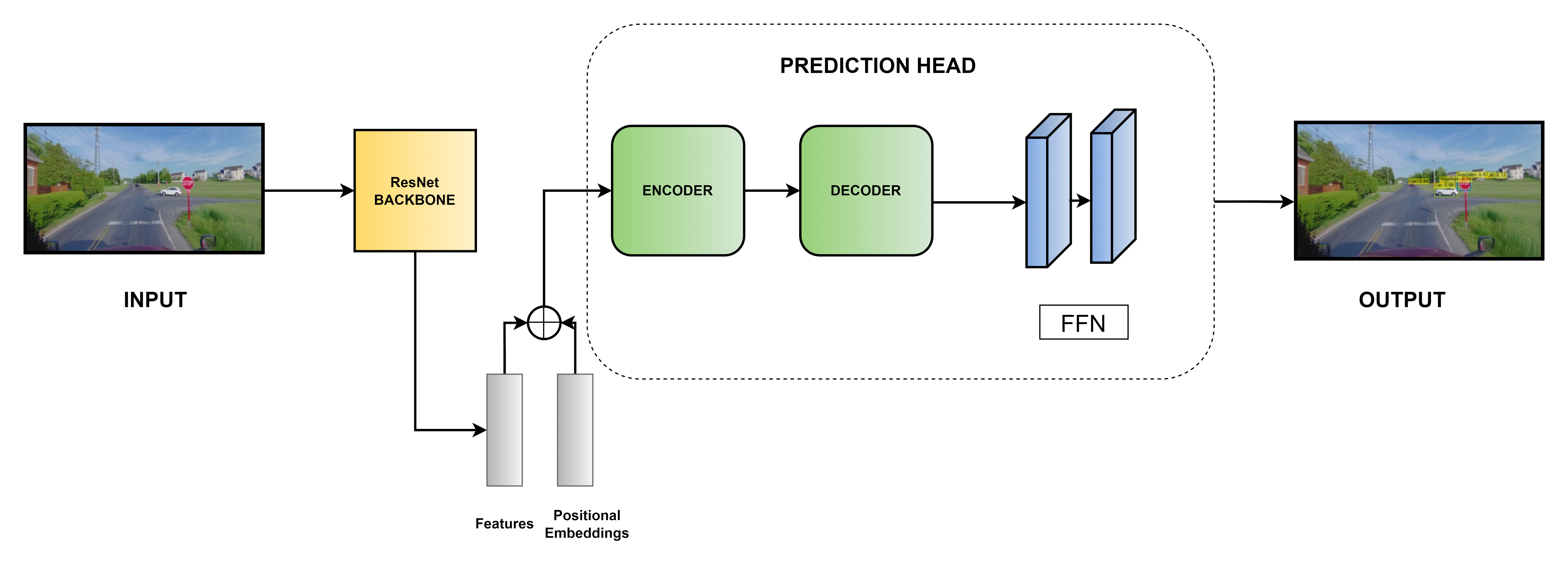}
          \label{Fig:System pipeline}
          \caption{System Pipeline}
\end{figure*}
\section{Dataset} \label{sec:dataset}
In this study, the dataset plays a key role. As discussed above, the major challenges in object detection are highly dynamic environments, truck blind spots, low light, and noisy environments. Thus we experimented with a competition dataset i.e Motive AI challenge dataset. The dataset contains 39,998 training images and 4001 validation images along with annotations respectively. There are four classes: traffic signals, stop signs, cars, and trucks. Motive is a leading fleet management company working on intelligent automation in trucking. Often, it has been observed that object detectors perform ideally on synthetic or self-curated datasets but performance drops significantly in real-world conditions. Thus this dataset has been prepared by considering the special focus on real-world conditions. By employing a dashcam on a fleet of trucks, this dataset has been collected in different day-lights and different weather. different scenes and different traffic conditions. Objects of interest i.e traffic signals, stop signs, cars, trucks and others are captured from different angles and positions in order to achieve generalization.

Figure \ref{Fig:dataset} display the sample images of dataset and for better understanding of the problem  a, b, c, d. These images represent the different on-road conditions covered by the dataset i.e. low-illumination, bad-weather, long-proximity and short-proximity objects. The dataset is designed such that it is ensures that performance does not drop in real-world conditions and the system generalizes well to most of the possible conditions in deployment.

\section{Methodology}\label{sec:method}
In this work, Transformer based detection architecture i.e DEtection TRansformer (DETR) is utilized for the task of object detection in imagery from the dashcams. This work is focused on the detection of objects from a dashcam point of view. There are many challenges in the case of object detection in intelligent dashcams. Some of the challenges are the dynamic environment on the road, the large number of vehicles on road, the large size of trucks as compared to the small size of other vehicles. We have seen in recent work, architectures such as RCNN and YOLO perform really well on object detection problems but they do not consider much contextual information while making decisions and it has been observed that contextual information is the key point in efficient decision-making in such highly dynamic environments \cite{ma2021miti}. Transformers-based architectures have been the best performing in Natural Language Processing (NLP) problems for a long time as they are still considered a key milestone by introducing self-attention-based decision-making. Recently they have also shown outstanding benchmark performances in the field of vision such as DETR, DINO, SWIN-Transformer and Vanilla ViT. Thus we have employed a DEtection TRansformer(DETR) to perform object detection in this research. 
\begin{figure}[h]
    \centering
    \includegraphics[width = 8cm, height=8cm]{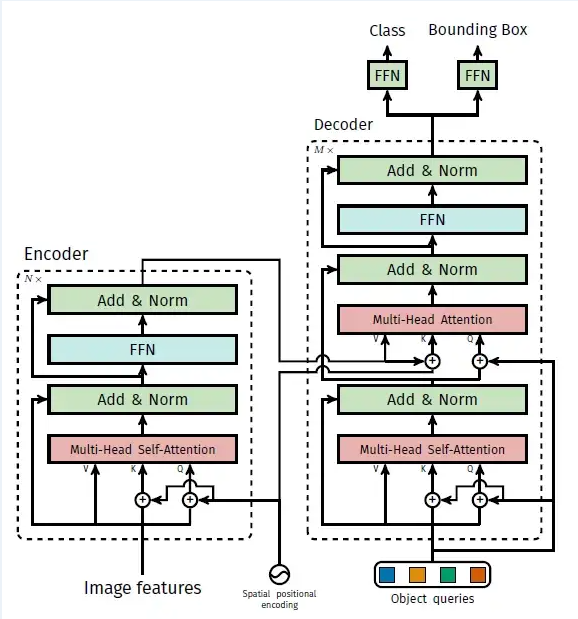}
    \caption{Detailed Architecture of Encoder-Decoder Block}
    \label{darc}
\end{figure}
\subsubsection{Architecture}
The network architecture is such that a transformer as prediction head on top of ResNet-50 backbone for feature extraction. Figure \ref{Fig:System pipeline} illustrates the pipeline of our proposed solution. The input image passes through a ResNet which performs the feature extraction. Features along with positional embeddings are passed into an encoder-decoder respectively. Figure \ref{darc} further details out the architecture of encoder-decoder. An encoder starts with a multi-head self attention followed by an add \& norm layer which connects with a FFN and finally an Add \& Norm layer is repeated. The decoder architecture follows a similar pattern with two FFN's on the head for final prediction, one for bounding box and one for class. The embedding size and no of layers for encoder-decoder is set to value of 256. Total 6 encoder-decoder layers have been set with 8 self-attention layers.

The encoder takes in the input sequence and produces an output which is then used by the decoder. The decoder also receives object queries which assist in the decoding of positional embeddings. Initially, these object queries are randomly generated vectors, but they are adjusted during the training process. The bipartite matching loss has been followed by the network for the computation of loss based on ground truth and predictions. The output of the decoder is then passed through a feedforward neural network, with the number of networks being equal to the number of object classes and each containing multiple layers.

Let us denote by $y$ the ground truth set of objects, and $\hat{y}=\left\{\hat{y}_i\right\}_{i=1}^N$ the set of $N$ predictions. Assuming $N$ is larger than the number of objects in the image, we consider $y$ also as a set of size $N$ padded with $\varnothing$ (no object). To find a bipartite matching between these two sets we search for a permutation of $N$ elements $\sigma \in \mathfrak{S}_N$ with the lowest cost. The bipartite matching loss:
\begin{equation}
\hat{\sigma}=\underset{\sigma \in \mathfrak{S}_N}{\arg \min } \sum_i^N \mathcal{L}_{\text {match }}\left(y_i, \hat{y}_{\sigma(i)}\right)
\end{equation}
where $\mathcal{L}_{\text {match }}\left(y_i, \hat{y}_{\sigma(i)}\right)$ is a pair-wise matching cost between ground truth $y_i$ and a prediction with index $\sigma(i)$. 
\begin{figure*}[!h]
\centering
\begin{tabular}{cc}
  \includegraphics[width=55mm,frame]{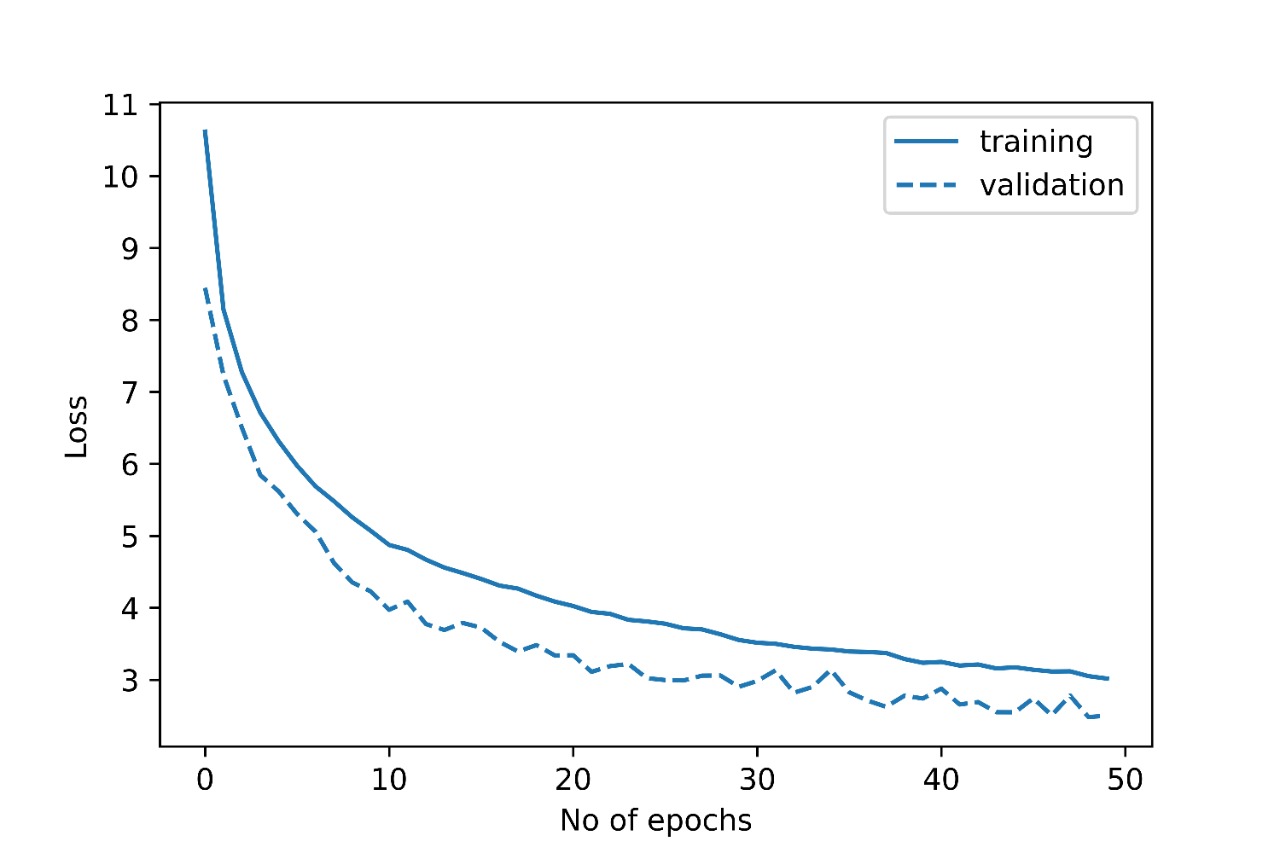} &   \includegraphics[width=55mm, frame]{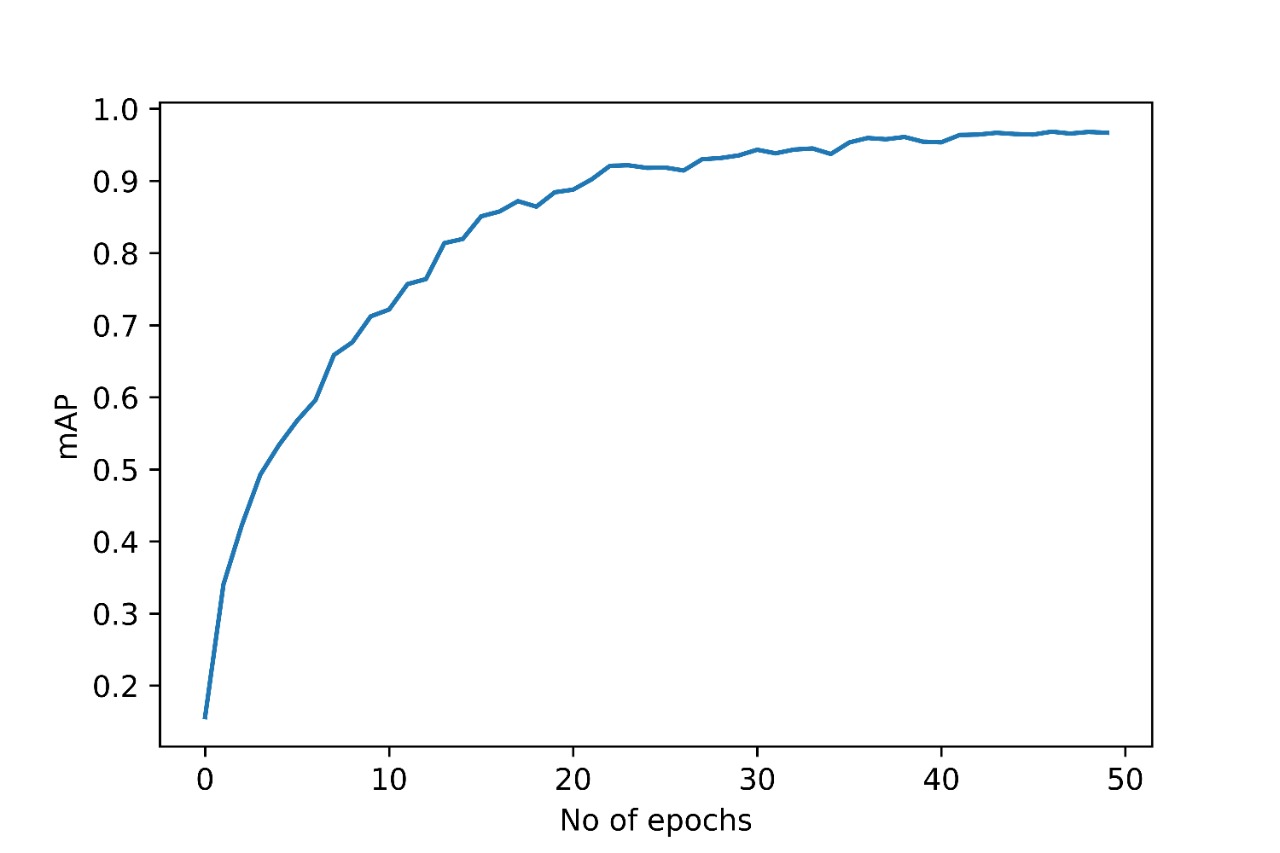} \\
  \label{fig:curve}
(a) Loss Curve & (b) mAP Curve\\[6pt]

\end{tabular}
\caption{Training Loss and mAP Curves }
\end{figure*}
\begin{figure*}[!ht]
\centering
\begin{tabular}{cc}
  \includegraphics[width=60mm]{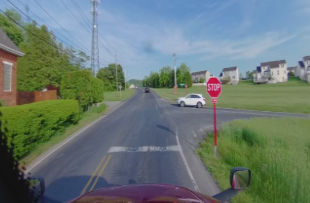} &   \includegraphics[width=60mm]{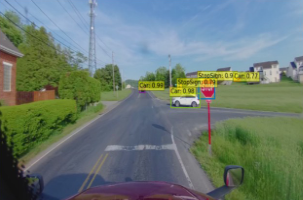} \\
(a) & (b) \\[6pt]
\end{tabular}
\caption{Inference Results}
\label{fig:visualization results}
\end{figure*}

\subsubsection{Training Configuration}
We performed the training configuration as illustrated in Table~\ref{tab:my-tableTC} . The total training time is 45 hours.

\begin{table}[!ht]
\centering
\caption{Train Configuration}
\label{tab:my-tableTC}
\begin{tabular}{ll}
\hline
\textbf{Parameter} & \textbf{Value}    \\ 
\hline
No of Parameters   & 41.3M             \\
GPU                & NVIDIA Tesla P100 \\
Epochs             & 50                \\
learning-rate (lr)                 & 1e-5              \\
lr-backbone        & 1e-6              \\
Optimize           & Adam              \\
batch-size         & 8                 \\
num-queries        & 100               \\
\hline
\end{tabular}
\end{table}

\section{Experiments and Results} \label{sec:res}
In this section, we present the details of experimentation carried out in order to perform quantified validation of our proposed methodology. This section will be used as a reference to reproduce the results anytime. Details of the dataset, evaluation metrics and experimentation are provided further in this section respectively.

\subsection{Evaluation Metrics}
As this is an object detection problem, the basic evaluation metrics are Precision and Recall used in to evaluate the detection performance. Precision and recall are calculated using True Positive (TP), False Positive (FP), True Negative (TN) and False Negative (FN). 


Precision and Recall are calculated using:
\begin{equation}    
Precision = \frac{TP}{TP+FP}
\end{equation}
\begin{equation}
Recall = \frac{TP}{TP+FN}
\end{equation}
The average precision and recall for each class is calculated, and the mean Average Precision (mAP) and mean Average Recall (mAR) is determined at different IoU values as follows:

mAP50 is computed by averaging the mAP using a 50
\begin{enumerate} 
  \item {$\mathbf{mAP}$}The mAP is calculated by taking the average of its values at 10 different IoU thresholds, which range from 50\% to 95\%
  \item {$\mathbf{mAP_{50}}$} is calculated by taking the average mAP over 50\% IoU threshold.
  \item {$\mathbf{mAR_{10d}}$} is calculated by the maximum recall values given 10 detections per image by taking average over IoUs and all the classes..
  \item {$\mathbf{mAR_{100d}}$} is calculated by the maximum recall values given 100 detections per image taking average over IoUs and all the classes.
\end{enumerate}
\subsection{Results and Analysis}
This section details the experimental results and analysis. DETR fine-tuned on our custom dataset achieves an average mAP of 0.95 with the IOU threshold set to 0.50 as shown in Table~\ref{tab:DETR_4}. It is observed that this mAP is a great result considering the challenging conditions covered in the dataset. It can be seen in Figure \ref{fig:curve} that the mAP is improving as the epochs increase which shows smooth learning. Considering the different harsh real-world conditions such as low light, noisy environments, and occluded and cluttered environments, the results are quite impressive. As this is a private and proprietary dataset there is no related work for comparison.

Figure \ref{fig:visualization results} displays inference results, images on the left are actual inputs and on the right are output images from DETR. In the first row, it can be seen that the network accurately detects and objects of interest. In the first row, two bounding boxes are detected, one with a confidence of 0.90 and the other with a confidence of  0.79. The one with confidence 0.79 is rarely visible if we just consider its visual characteristics, it would be extremely difficult to detect it if a detector just considers its visual features as the "stop-sign" text is completely not visible, but DETR has efficiently detected this stop sign with solid confidence by considering global contextual information.
Furthermore, in the second row, it can be seen that it is a low-light environment, you can check by zooming in that even a human would find it extremely difficult to detect many cars parked alongside left and right on the road in the parking of a gas station. DETR performs outstandingly in this condition and detects all the vehicles parked in dark parking. If a detector just considers the visual features in this image, the darkness is more dominant and it would not be able to detect all those cars, but DETR being a transformer has efficiently detected by considering global contextual information. These experimental results validate that state-of-the-art DETR can be a robust object detection network for truck dashcams in harsh conditions.  
\begin{table}[h!]
\setlength{\tabcolsep}{12pt}
\centering
\caption{Detection Results of DETR (Detection transformer) 
\label{tab:DETR_4}}
\begin{tabular}{lp{1.5cm}p{1.0cm}p{1.0cm}p{1.0cm}}
\hline
\textbf{Metrics}   & \textbf{Accuracy} \\ \hline
 $\mathbf{mAP}$  &    0.623     \\ 
 $\mathbf{mAP_{50}}$    &    0.951    \\ 
 $\mathbf{mAR_{10d}}$  & 0.703   \\ 
$\mathbf{mAR_{100d}}$    &   0.726  \\ \hline
 
\end{tabular}
\end{table}

\section{Conclusion} 
\label{sec:con}
In this study we have proposed a transformers based object detection solution for the challenging problem of  object detection in vehicle dashcams in a highly dynamic environment. We have fine-tuned a state-of-the-art DEtection TRansformer (DETR) for this purpose on our custom dataset. The experimental results discussed in results and analysis section validates that DETR is a robust network for performing object detection in harsh deployment conditions such as low levels of illumination due to to weather and different daytime, occluded and noisy environment and highly dynamic environments. The point that in this proposed solution, network considers contextual information makes it robust and an efficient solution for this highly challenging problem. This study validates that if DETR is trained on a generalized data distribution covering major deployment conditions DETR generalizes really good in deployment and the performance does not drops. This study can be utilized as one of the references for future work in this domain.

\bibliographystyle{ieeetr}
\bibliography{Reference}

\end{document}